\documentclass[sigconf, authorversion]{acmart}

\usepackage{cleveref}
\usepackage{tabularx}
\usepackage{multirow}
\usepackage{enumitem}
\usepackage{graphicx} 
\usepackage{booktabs} 
\usepackage{float}    
\usepackage{subcaption} 

\setcopyright{acmlicensed}
\copyrightyear{2025}
\acmYear{2025}
\acmDOI{XXXXXXX.XXXXXXX}

\acmConference[WebSci '25]{ACM Web Science Conference}{May 2025}{Online}
\acmBooktitle{Proceedings of the ACM Web Science Conference (WebSci '25), May 2025}
\acmISBN{978-1-4503-XXXX-X/2025/05}

\title{Early Multimodal Prediction of Cross-Lingual Meme Virality on Reddit: A Time-Window Analysis}

\author{Sedat Dogan}
\email{s.dogan-2021@hull.ac.uk}
\orcid{0009-0002-7409-7696}
\affiliation{%
  \institution{School of Computer Science, University of Hull}
  \city{Hull}
  \country{UK}
}

\author{Nina Dethlefs}
\email{n.dethlefs@lboro.ac.uk}
\affiliation{%
  \institution{Loughborough University}
  \city{Loughborough}
  \country{UK}
}

\author{Debarati Chakraborty}
\email{D.Chakraborty@hull.ac.uk}
\affiliation{%
  \institution{School of Computer Science, University of Hull}
  \city{Hull}
  \country{UK}
}

\renewcommand{\shortauthors}{Dogan et al., 2026}

\begin{document}

\begin{abstract}
Memes are a central part of online culture, yet their virality remains difficult to predict, especially in cross-lingual settings. We present a large-scale, time-series dataset of \textbf{46{,}578} Reddit memes collected from \textbf{25} meme-centric subreddits across \textbf{eight} language groups, with more than \textbf{one million} engagement tracking points. We propose a data-driven definition of virality based on a Hybrid Score that normalises engagement by community size and integrates dynamic features such as velocity and acceleration. This approach directly addresses the field’s reliance on static, simple volume-based thresholds with arbitrary cut-offs.

Building on this target, we construct a multimodal feature set that combines Visual, Textual, Contextual, Network, and Temporal signals, including structured annotations from a multimodal LLM to scale cross-lingual content labelling in a consistent way. We benchmark interpretable baselines (XGBoost, MLP) against end-to-end deep models (BERT, InceptionV3, CLIP) across early observation windows from \textbf{30} to \textbf{420} minutes. Our best model, a multimodal XGBoost classifier, achieves a PR AUC of \textbf{0.43} at \textbf{30} minutes and \textbf{0.80} at \textbf{420} minutes, indicating that early prediction of meme virality is feasible even under strong class imbalance.

The results reveal a clear \emph{Content Ceiling}, where content-only and deep multimodal baselines plateau at low PR AUC, while \textbf{structural Network and Temporal features are necessary to surpass this limit}. A SHAP-based temporal analysis further uncovers an \emph{evidentiary transition}, where early predictions are dominated by network priors (author and community context), and later predictions increasingly rely on temporal dynamics (velocity, acceleration) as engagement accumulates. Overall, we \textbf{reframe meme virality as a dynamic, path-dependent process governed by exposure and early interaction patterns rather than by intrinsic content alone}.
\end{abstract}



\maketitle

\section{Introduction}
Online platforms host a constant flow of content, and memes often act as a primary mode of communication, cultural commentary, and information spread\cite{shifman_cultural_2014}. The rapid replication and wide reach of memes is often described metaphorically as viral, similar to the spread of biological viruses but driven by psychological and emotional mechanisms rather than biological ones\cite{denisova_how_2020}. Understanding which memes will achieve widespread popularity, or ``go viral'', is a complex but important problem with implications for social media platforms, marketers, and researchers studying the dynamics of information diffusion.

Forecasting virality early in a meme's lifecycle (within minutes or hours) is particularly challenging. Sparse initial data and noisy early signals make prediction difficult, yet early forecasts are highly valuable for proactive content moderation, improved recommendation, and timely analysis of emerging trends. Memes often require culturally specific knowledge to decode humour and meaning, with interpretations that vary significantly between cultural groups\cite{guseynova_reflection_2022}. Previous research has explored virality prediction using a range of features and models, including deep learning techniques\cite{chen_npp_2019}. Despite this, several gaps remain. Many studies focus on a single platform or language, adopt heterogeneous and often arbitrary definitions of virality, and lack a granular analysis of how predictive power evolves during the crucial early hours. Although there are specific studies on meme virality\cite{barnes_dank_2020, ling_dissecting_2021, sah2025decoding}, \textbf{the establishment of a robust, data-driven definition of virality and benchmarks for early prediction in diverse communities remains an open area}.

\begin{figure*}[htbp]
    \centering
    \includegraphics[width=0.9\textwidth]{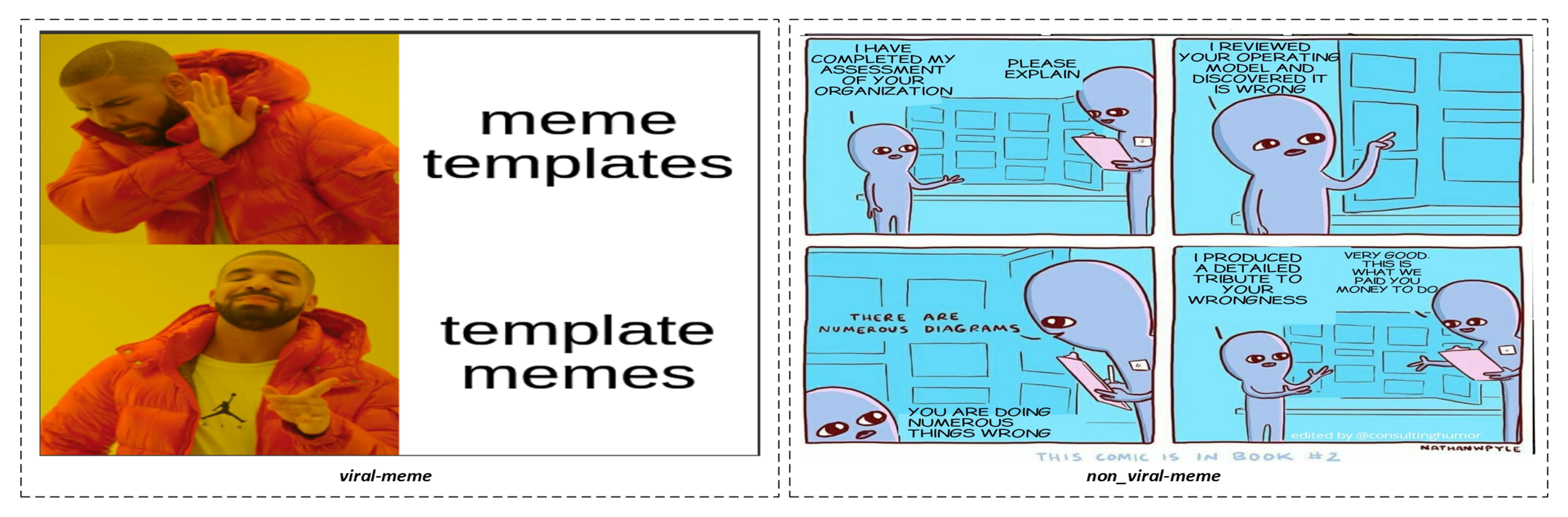} 
    \caption{Examples of memes classified as viral (left, Drake meme template) and non-viral (right, multi-panel comic meme) under our data-driven definition.}
    \label{fig:meme_examples} 
\end{figure*}

We aim to bridge these gaps and investigate the following research questions.
\begin{enumerate}
    \item Can we establish a robust, data-driven definition of virality based on a hybrid engagement score that avoids arbitrary volume thresholds?
    \item How early can we accurately predict meme virality using combined engagement dynamics, network context, and static content features?
    \item How does the performance of the predictive models change as the observation window lengthens, and what is the relative importance of different feature categories for early prediction? How does this importance shift over time?
\end{enumerate}

We hypothesise that signals of meme virality are dynamic and follow a phased progression that we call an ``evidentiary transition''. In this view, the nature of predictive evidence shifts from static context to dynamic engagement as a meme's lifecycle unfolds.

To address these questions, we use a large-scale, cross-lingual dataset from \textbf{25} Reddit communities. We propose and apply a method for defining virality based on a hybrid score of weighted engagement and dynamic features. We engineer a comprehensive multimodal feature set including LLM-derived semantic tags as a pragmatic alternative to manual annotation, trading token-level supervision for scalable, cross-lingual consistency. We systematically evaluate a range of models, including XGBoost, across multiple time windows from \textbf{30} to \textbf{420} minutes, alongside deep baselines such as BERT and CLIP.

Our results show that this feature-rich approach, when paired with XGBoost, \textbf{outperforms deep learning baselines in both predictive performance and computational efficiency}. Our analysis \textbf{confirms the evidentiary transition hypothesis}, revealing a clear temporal shift in the importance of features from initial network and contextual signals to observed engagement dynamics. \cref{fig:meme_examples} illustrates two meme examples that our model classifies as viral and non-viral.

\section{Related Work} \label{sec:related_work}
The concept of ``virality'' has become a complex, interdisciplinary construct that reflects shifts in digital communication, network structures, and cultural dynamics. Memes propagate differently across cultures and languages, shaped by local socio-cultural contexts, linguistic nuances, and visual traditions \cite{ageev_cognitive_2024}. In this section, we review work on (i) how virality is defined and operationalised on social media, (ii) multimodal and contextual predictors of meme success, and (iii) temporal models of online diffusion.

\subsection{Defining and Predicting Meme Virality on Social Media}
In media and communication studies, virality is a multifaceted concept that goes beyond simple popularity. It is usually defined as the rapid and wide spread of content through decentralised digital networks, where speed and reach distinguish it from traditional word of mouth communication \cite{mills_virality_2012}. This spread is not purely mechanical. Unlike biological contagion, media virality is driven by users' emotional and psychological involvement, which provides the main motivation to share \cite{denisova_how_2020}.

Defining meme virality in social media research still relies heavily on volume-based thresholds applied to static final engagement counts. A common approach frames virality as a binary classification task using arbitrary percentiles. For instance, \cite{barnes_dank_2020} and \cite{barnes_topicality_2024} label the top \textbf{5\%} of posts by normalised upvotes as viral, with the latter exploring thresholds between \textbf{1\%} and \textbf{25\%}. To sharpen the distinction between classes, \cite{sah2025decoding} compare the top \textbf{5\%} with the bottom \textbf{5\%} and exclude the middle majority.  Other work uses multi-metric criteria or repetition frequency. \cite{elamrany2025redditv} define virality in the Reddit-V dataset as posts that fall in the top \textbf{20\%} for both upvotes and comment counts at the same time. In the context of image propagation on 4chan, \cite{ling_dissecting_2021} measure virality by image repetition, contrasting the top \textbf{100} most posted images with the bottom \textbf{100}. Despite these variations, most studies still ground success in static, ex-post metrics rather than dynamic temporal properties.

\subsection{Multimodal Meme Virality Prediction}
The literature traces a clear path from theory to empirical models of virality. We view virality as the result of interactions between network structure, content features, and human behaviour. Researchers have moved beyond raw engagement counts toward measures that capture structural and community-relative diffusion \cite{goel_structural_2015, weng_virality_2013}. For memes, studies show that combining visual and textual signals outperforms single-modality approaches. Methodologically, the field still balances high-performing black-box models against more interpretable alternatives. Recent work explores multimodal prediction by combining visual and textual features with machine learning models such as Random Forests and CNNs. For instance, \cite{sah2025decoding} report approximately \textbf{80\%} accuracy on English memes and find that specific visual attributes, such as lighter colours, multiple objects, and positive sentiment, are positively correlated with high engagement. Similarly, \cite{ling_dissecting_2021} report a high ROC-AUC of \textbf{0.86} on 4chan data and note that close-up shots and distinct emotional expressions are strong predictors of success.

However, the sufficiency of content features alone remains debated \cite{barnes_dank_2020}. To address this limitation, recent research uses Large Language Models (LLMs) as zero-shot reasoners, with chain-of-thought prompting to decompose multimodal tasks without specific training. While this approach reduces annotation costs and scales well, it also introduces challenges \cite{xu_casflow_2023}. LLMs are sensitive to prompt design and prone to factual errors \cite{wang_agent_2023}. Risks around bias, task contamination, and data privacy \cite{liu_case-based_2024} further show that, although powerful, LLM-based extraction needs careful engineering and critical evaluation. In this work, we treat a multimodal LLM as an annotation tool rather than a predictor, using it to convert raw content into structured labels that can be inspected and audited like any other feature.

Many studies show that combining modalities helps, yet few models integrate \textbf{all three feature families} content, network, and temporal dynamics with fine-grained measures such as velocity and acceleration for early prediction.

\subsection{Temporal Modelling of Online Content Virality}
Predicting online content virality shortly after publication is difficult yet crucial for proactive content management. It requires models that can capture complex dynamics through different modelling frameworks. Self-exciting point processes such as Hawkes processes are widely used to model how past engagements increase the likelihood of future interactions \cite{zhang_survey_2021}. More recently, graph-based deep learning has gained prominence, with Graph Neural Networks (GNNs) and Transformers such as Casformer using attention mechanisms or topological analysis to capture structural and temporal features at the same time \cite{zhang_casformer_2025}. Some advanced models also employ Neural ODEs to handle continuous-time irregularity \cite{cheng_information_2024}. 

We draw a key distinction between continuous-time and discrete-time modelling. Continuous-time models handle asynchronous, high-frequency data particularly well \cite{cieslewski_continuous-time_2022, de_haan-rietdijk_discrete-_2017}. Discrete-time approaches can lose detailed timing information or introduce bias \cite{de_haan-rietdijk_discrete-_2017}, but they reduce implementation complexity and make feature comparison over fixed intervals more straightforward. We therefore adopt a discrete-time framework as a practical way to analyse how predictive signals evolve across stages of the meme lifecycle.  

Many graph-based methods also require complete knowledge of the diffusion cascade, which is not available in real time. We address this by using intrinsic content features and aggregate engagement dynamics, which are readily accessible and offer greater interpretability than black-box GNNs. Within our \textbf{discrete-time} setup, and despite the potential information loss noted in \cite{gleeson_limitations_2016, de_haan-rietdijk_discrete-_2017}, these features provide a practical and effective basis for analysing how predictive signals change over the content lifecycle.

\subsection{Research Gaps and Our Contribution}
This review highlights several gaps that our work aims to address. First, there is a dataset gap in both temporal resolution and linguistic diversity. Second, in terms of multimodal integration, current methods often overlook dynamic temporal and network features and focus mainly on static content properties. Third, while temporal dynamics are known to matter for cascade predcition, systematic analyses of how feature importance shifts across early time windows are still rare for meme virality prediction.Finally, many studies rely on simple, arbitrarily chosen thresholds to define virality and ignore its dynamic nature in their experimental design, making results difficult to compare between studies. 

Our study makes several contributions to address these gaps. We introduce a \textbf{large-scale, cross-lingual dataset with high-resolution temporal data}. We tackle the definitional challenge by proposing and implementing a \textbf{data-driven methodology to define virality} that is designed to be both robust and methodologically sound. We then systematically evaluate a range of models, from interpretable methods to strong deep neural networks, using a comprehensive static and dynamic multimodal feature set, where the static content and contextual features are derived via LLM-based zero-shot extraction. Finally, we move beyond simple prediction and \textbf{analyse the temporal evolution of predictive signals}.

\section{Data and Methodology} \label{sec:data_methodology}

This section describes our data collection, preprocessing, and predictive modelling framework. We place strong emphasis on avoiding data leakage to keep our results valid.

\subsection{Dataset Construction and Scope} \label{subsec:dataset_construction}
A dataset that captures the dynamics of cross-lingual memes forms the foundation of this research. We collected this dataset using the official Reddit API (PRAW) between \textbf{21 March and 31 May 2025}. To ensure diversity, we collected data from \textbf{25} distinct meme-centric subreddits across \textbf{eight} language groups. Table~\ref{tab:media_dist} shows the distribution of media types between language groups after filtering. For each meme post ($j$), we collect standard metadata (e.g., title, anonymised author, timestamp, URL of the media) and subreddit context (subscriber count $N_i$). 

\begin{table}[htbp]
    \centering
    \caption{Distribution of media types across language groups. Note: counts represent the distribution after the final filtering step (e.g., requiring $> 24$ h tracking, valid media).}
    \label{tab:media_dist}
    \small 
    \begin{tabular}{@{}lrrrrrr@{}}
        \toprule
        Language   & Image & Video & GIF & Text & Audio & Total \\
        \midrule
        English    & 35,206 & 2,404  & 797 & 2    & 1     & 38,410 \\
        German     & 3,825  & 72    & 16  & 0    & 1     & 3,914  \\
        Mixed      & 1,555  & 135   & 38  & 2    & 1     & 1,731  \\
        Nordic     & 699   & 45    & 3   & 0    & 0     & 747   \\
        Spanish    & 516   & 200   & 9   & 0    & 1     & 726   \\
        French     & 505   & 36    & 10  & 0    & 0     & 551   \\
        Portuguese & 170   & 44    & 7   & 0    & 0     & 221   \\
        Turkish    & 120   & 38    & 7   & 0    & 0     & 165   \\
        Italian    & 65    & 43    & 5   & 0    & 0     & 113    \\
        \midrule
        Total      & 42,661 & 3,017  & 892 & 4    & 4     & 46,578 \\
        \bottomrule
    \end{tabular}
\end{table}
A key element of the dataset is tracking engagement metrics, including score $s(t)$, comments $c(t)$, and crossposts $x(t)$. We track these with \textbf{high temporal resolution} using dynamic sampling, starting with \textbf{5}-minute intervals and decreasing the sampling frequency later. This strategy allows us to follow each meme's life trajectory and its spread across Reddit's internal ranking categories, which is crucial for early prediction, while also capturing longer-term trends. The final recorded engagement metrics inform the definition of the target variable. Descriptive statistics of the metadata are provided in Table~\ref{tab:dataset_metadata}.

\begin{table}[h!]
    \centering
    \small
    \renewcommand{\arraystretch}{1.3}
    \caption{Descriptive statistics of the dataset.}
    \label{tab:dataset_metadata}
    \begin{tabularx}{0.8\columnwidth}{X r}
        \toprule
        \textbf{Metadata Metrics} & \textbf{Count Value} \\
        \midrule
        \multicolumn{2}{l}{\textit{\textbf{Collection Scope}}} \\
        Observation Period & Mar 21, 2025 -- May 31, 2025 \\
        Total Unique Posts & 71,040 \\
        Total Time-Series Data Points & 1,465,958 \\
        Distinct Subreddits & 25 \\
        \addlinespace
        
        \multicolumn{2}{l}{\textit{\textbf{Temporal Resolution}}} \\
        Avg. Data Points per Post & 20.6 \\
        Data Points Distribution (Top 25\%) & $>$ 30 observations \\
        Tracking Interval (Avg) & $\approx$ 32 minutes (1,919 s) \\
        \addlinespace
        
        \multicolumn{2}{l}{\textit{\textbf{Top 5 Subreddits (by Volume)}}} \\
        r/memes & 15,375 \\
        r/shitposting & 9,766 \\
        r/formuladank & 9,753 \\
        r/antimeme & 5,929 \\
        r/ich\_iel (German) & 3,719 \\
        \addlinespace
        
        \multicolumn{2}{l}{\textit{\textbf{Author Engagement}}} \\
        Avg. Author Account Age & $\approx$ 3 years (94.8M s) \\
        Avg. Author Post Karma & 72,055 \\
        Avg. Author Comment Karma & 23,882 \\
        \bottomrule
    \end{tabularx}
\end{table}

Standard data cleaning removed moderated posts, posts without a valid URL, and posts without at least \textbf{24 hours} of continuous engagement tracking. Our process complied with Reddit's Terms of Use, preserved user anonymity, and received faculty approval for responsible conduct. In total, we initially collected \textbf{71{,}040} unique meme posts through the Reddit API. After applying quality filters (removing moderated posts, posts without a media link, and posts with less than \textbf{24 hours} of continuous tracking), our final dataset contains \textbf{46{,}578} unique meme posts with more than \textbf{one million} tracking points.

\subsubsection*{\textbf{Data Splitting and Experimental Integrity}} 
To ensure that our models are evaluated on truly unseen data, we split this filtered dataset chronologically. All posts from \textbf{21 March to 15 May 2025} (approximately \textbf{80\%} of the data) form the training set, and posts from \textbf{16 May to 31 May 2025} (approximately \textbf{20\%}) make up the held-out test set (Table~\ref{tab:dataset_split}). This chronological split prevents data leakage and creates a realistic evaluation setting in which models are tested on future data relative to their training period.

\begin{table}[htbp]
    \centering
    \small 
    \setlength{\tabcolsep}{4pt} 
    \caption{Chronological split of the final filtered dataset.}
    \label{tab:dataset_split}
    \begin{tabular}{@{}lrrr@{}}
        \toprule
        \textbf{Split} & \textbf{Date Range} & \textbf{Count} & \textbf{Percent} \\
        \midrule
        Training Set & Mar 21--May 15, 2025 & 37,262 & 80.0\% \\
        Test Set & May 16--May 31, 2025 & 9,316 & 20.0\% \\
        \midrule
        \textbf{Total} & \textbf{Mar 21--May 31, 2025} & \textbf{46,578} & \textbf{100\%} \\
        \bottomrule
    \end{tabular}
\end{table}

\subsection{Target Variable Definition: A Data-Driven Approach} \label{subsec:target_definition}
To prevent data leakage, we define virality using only the training portion of the dataset and then apply this definition to the held-out test set. Instead of relying on arbitrary thresholds, we use a data-driven approach that reflects both the volume and dynamics of engagement. We perform the following steps only on the training data.

\begin{enumerate}
    \item \textbf{Normalisation by Community Size:} We normalise the raw engagement metrics ($k \in \{s, c, x\}$) for each post by its subreddit subscriber count $N_i$ to remove the impact of subreddit size on the engagement metrics. To reduce the impact of extreme values, we apply a \textbf{99th percentile} cap ($P_{99}$) computed from the training distribution:
    \begin{equation} \small 
        m_{j,k}(t) = \min\left(\frac{k_j(t)}{N_i} \times 100000, \, P_{99\_train}\left(\frac{k}{N} \times 100000\right)\right).
    \end{equation}

    \item \textbf{Hybrid Engagement Weighting:} To estimate the relative importance of different engagement signals, we train an auxiliary Random Forest model on the training set. This model uses both normalised volume metrics (score, comments, crossposts) and key dynamic features (e.g., velocity, acceleration, time to take-off) from early time windows (30, 60, 120, \ldots, 420 min) to predict a preliminary target (top \textbf{5\%} by unweighted final sum). We then average the feature importances across windows to obtain a set of hybrid weights ($\beta_k$) for each feature $k$. This data-driven step confirms that the normalised score ($\beta_{\text{score}} = \textbf{1.0}$) and comments ($\beta_{\text{comments}} = \textbf{0.45}$) are the strongest volume signals, while dynamic features such as peak velocity ($\beta_{\text{peak\_vel}} = \textbf{0.14}$) also contribute meaningfully.

    \item \textbf{Final Hybrid Score:} We compute a composite Hybrid Score ($HS_{j,\text{final}}$) for each post using its final engagement features ($f_{j,k}$) and the learned hybrid weights ($\beta_k$):
     \begin{equation} \small
         HS_{j,\text{final}} = \sum_{k \in \text{features}} \beta_k \cdot f_{j,k}.
     \end{equation}

    \item \textbf{Threshold Identification:} To define the viral class in an objective way, we apply K-Means clustering\cite{MacQueen1967} with \textbf{$k=2$} to the $HS_{j,\text{final}}$ distribution. The resulting boundary ($\tau \approx \textbf{320.38}$) serves as our data-driven threshold (see \cref{fig:wes_threshold}), providing a natural separation between viral and non-viral posts.

    \begin{figure}[htbp]
        \centering
        \includegraphics[width=\columnwidth]{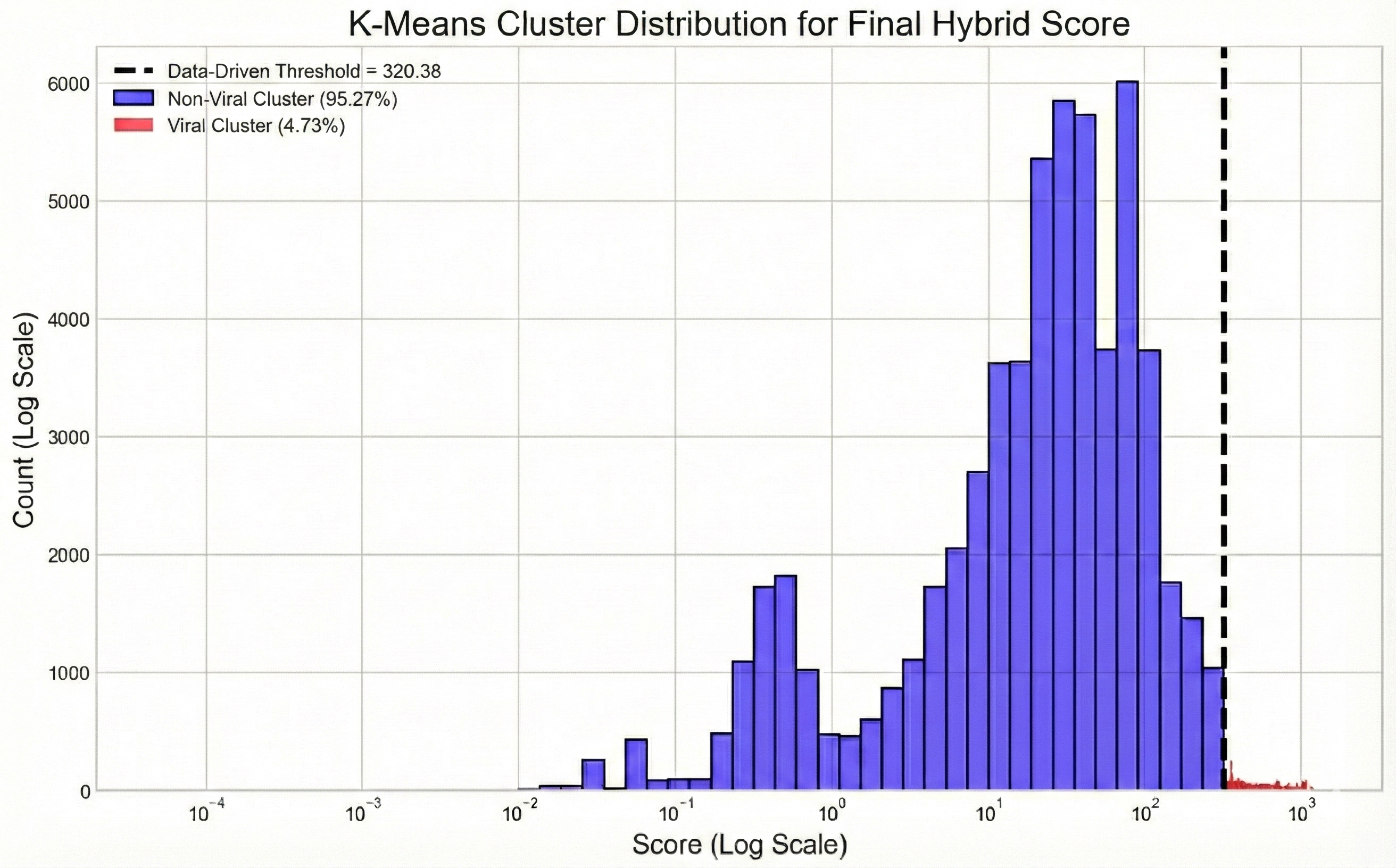}
        \caption{Distribution of final Hybrid Scores across training posts. The vertical line marks the data-driven threshold of \textbf{320.38} identified by K-Means, separating Non-Viral (\textbf{95.27\%}) and Viral (\textbf{4.73\%}) memes.}
        \label{fig:wes_threshold}
    \end{figure}

    \item \textbf{Final Target Variable Assignment:} We then use this single threshold, $\tau_{\text{train}}$, to assign the binary target \texttt{is\_viral} to \textbf{all posts in both the training and test sets}:
    \begin{equation}
        \texttt{is\_viral}_j = 
        \begin{cases}
            1 & \text{if } HS_{j,\text{final}} \ge \tau_{\text{train}}, \\
            0 & \text{if } HS_{j,\text{final}} < \tau_{\text{train}}.
        \end{cases}
    \end{equation}
\end{enumerate}

This procedure ensures that no information from the test set influences the virality definition, so our evaluation reflects a realistic, forward-looking prediction setting.
    
\subsection{Feature Engineering} \label{subsec:feature_engineering}
We engineered features that capture both dynamic engagement and static characteristics and ensured that all features for a time window $W$ use \textbf{only data available up to $W$}. All scaling and normalisation parameters were fitted on the training data and then applied to the test data. The final feature set is divided into several modality groups (Table~\ref{tab:feature_examples_updated}).

\begin{figure}[htbp]
    \centering
    \includegraphics[width=\columnwidth]{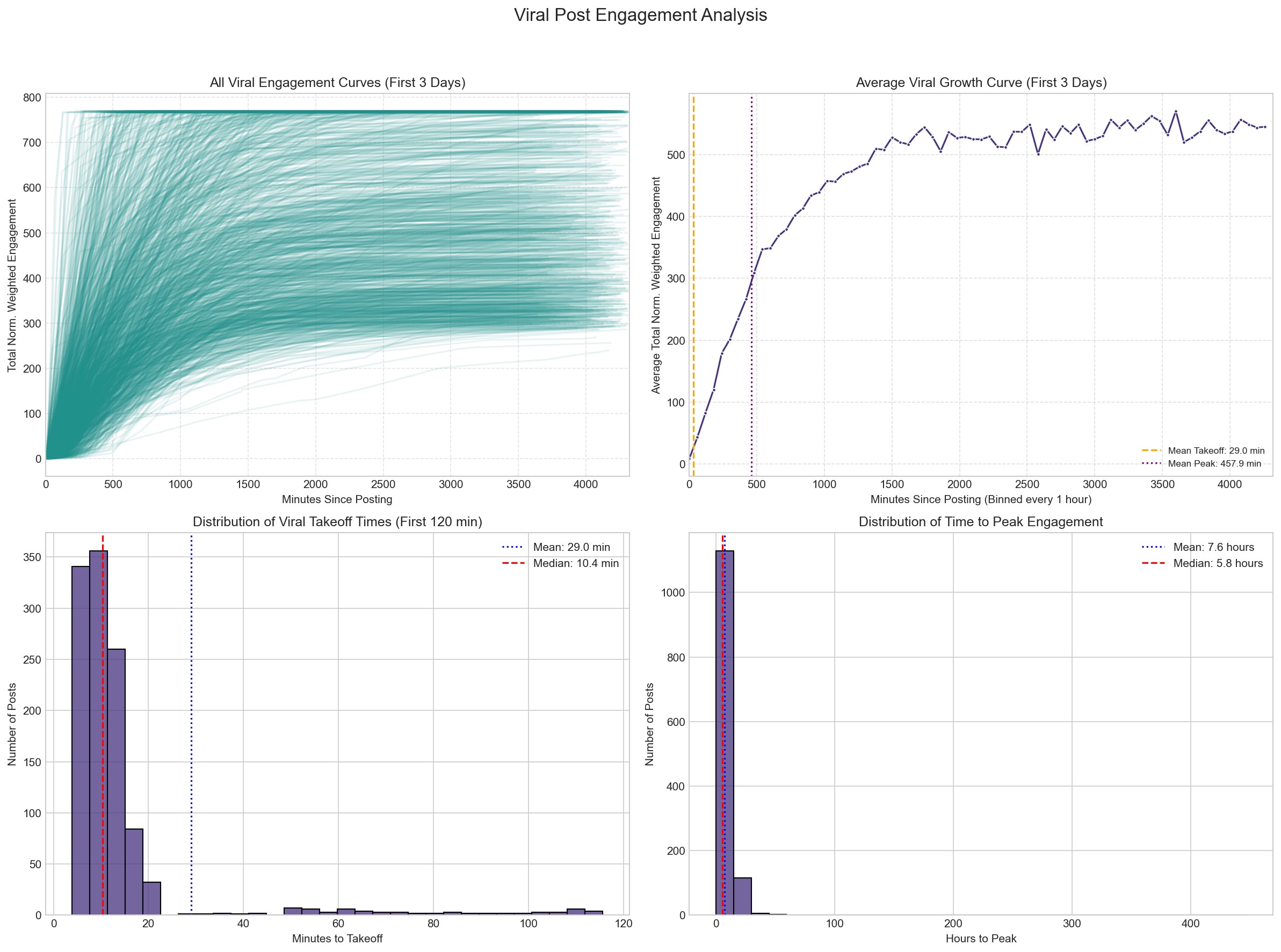}
    \caption{Lifespan trajectories of viral memes. Top-left: individual engagement curves for all viral posts. Top-right: average viral growth curve. Bottom-left: distribution of viral take-off times. Bottom-right: distribution of times to reach peak engagement velocity.}
    \label{fig:viral_trajectory}
\end{figure}

\begin{itemize}
    \item \textbf{Temporal Features.} We compute static temporal features (posting time/day) and dynamic metrics (velocity, acceleration) using strict causal windows. \cref{fig:viral_trajectory} shows the average dynamic trajectories of viral memes.
    \item \textbf{Network Features.} We compute dynamic network features (time in \textit{New}, category changes) and static network features (karma, subreddit size).
    \item \textbf{LLM-Derived Static Features.} For static content features, we use a multimodal LLM (Gemini\cite{GoogleGemini}) as a zero-shot annotation tool to extract structured visual, textual, and contextual features. Gemini is chosen for its multimodal, cross-lingual robustness and generous free daily API quota. In principle, the same pipeline could be recreated with strong open-source multimodal LLMs as they improve.
\end{itemize}

\begin{table}[htbp]
\small
\centering
\caption{Example features by modality.}
\label{tab:feature_examples_updated}
\begin{tabular}{@{}p{0.35\columnwidth}p{0.55\columnwidth}@{}}
\toprule
\textbf{Modality} & \textbf{Example feature(s)} \\
\midrule
Temporal & Posting time/day, velocity, acceleration, burst count, time to take-off/peak \\
Network & Category transitions, time in \textit{New}, fallback count, author post/comment karma, subreddit size \\
Visual & Objects, facial expression, composition, panels, video/audio duration \\
Textual & Text/title sentiment, text/title brevity, text language, text tone \\
Contextual & Meme template, target audience, offensiveness, topic \\
\bottomrule
\end{tabular}
\end{table}

\subsection{Exploratory Data Analysis (EDA)} \label{subsec:eda}
At the modality level, EDA shows that videos are much more likely to become viral than static images or GIFs (\textbf{16.4\%} vs.\ \textbf{6.0\%}), and minimalist layouts with human faces perform best. In the textual modality, longer captions and titles with nostalgic or broadly positive sentiment are associated with higher virality, and Nordic and German posts are more ``efficient'' than English posts given their smaller volume. Network indicators are particularly strong. Posts from smaller subreddits and posts that move from \textit{New} to \textit{Hot} show sharply elevated success rates. Temporally, early morning and mid-week posts (especially around \textbf{06:00} on \textbf{Tuesdays}) and rapid early engagement velocity within the first \textbf{20 minutes} are characteristic of viral memes Contextual labels indicate that offensive and politically or socially themed memes, and those targeted at specific fandoms, are more likely to go viral than neutral, broad-audience memes. (Appendix \cref{fig:visual_comps} and \cref{fig:cultural_audience}).

To examine cross-lingual differences among viral memes, we analyse the distribution of visual, textual, and contextual labels by language group. Visually, all communities are dominated by static images, but Turkish subreddits contain a higher share of videos, and memes with faces are especially common in Nordic and French posts. Textually, neutral sentiment and ironic or sarcastic tone dominate across languages, with Spanish and Portuguese memes showing more positive or humorous tone, and Portuguese and Turkish memes tending to use longer captions. Contextual patterns show that template usage is high in Nordic and mixed-language communities but lowest in French subreddits. Most memes target a general or Gen~Z/millennial audience, while English, German, and mixed-language subreddits contribute more niche or fandom-specific content. Among memes flagged as offensive, the mix of offence types varies by language group, with hate-speech and offensive-language categories relatively more prominent in Nordic and mixed-language communities, and sexual or relationship-related content dominating in some Romance-language subreddits. (Appendix \cref{fig:textual_crosslingual} and \cref{fig:contextual_crosslingual}).  Overall, these cross-lingual distributions confirm that surface content and audience profiles are culturally specific, while later sections show that structural (Network) and dynamic (Temporal) features provide the most stable and transferable signal for predicting virality

\subsection{Experimental Setup} \label{sec:experimental_setup}

This section describes our experimental setup, including data preprocessing, the models we evaluate, and the metrics we use.

\textbf{Preprocessing.}  
We implement a standard preprocessing pipeline using \texttt{scikit-learn}'s \texttt{Pipeline} and \texttt{ColumnTransformer} utilities. Numerical features are imputed with the median and then scaled with \texttt{StandardScaler}, while categorical features are one-hot encoded with missing values mapped to a constant \texttt{`missing'} category\cite{pedregosa2011sklearn}. \textbf{All preprocessing parameters are learned from the training set only and then applied to the test set}.

\textbf{Handling Class Imbalance.}  
Given the strong class imbalance in virality prediction (\textbf{4.73\%} viral posts), we address this by using class weights in our models to increase the penalty for misclassifying the minority (viral) class.

\textbf{Models Evaluated.}  
We evaluate a set of models that span interpretable linear baselines and deep learning architectures.
\begin{itemize}
    \item \textbf{XGBoost.} A gradient boosting framework that serves as our primary tree-based baseline and is well suited to tabular data\cite{chen2016xgboost}.
    \item \textbf{MLP Neural Network.} A standard Multi-layer Perceptron with two hidden layers that provides a deep learning baseline for tabular features\cite{Rumelhart1986}.
    \item \textbf{Deep Learning Baselines.} To benchmark our feature-based approach against end-to-end representation learning, we also evaluate \textbf{BERT} (Textual)\cite{devlin2019bert}, \textbf{InceptionV3} (Visual)\cite{szegedy2016rethinking}, and \textbf{CLIP} (Multimodal)\cite{Radford2021CLIP}.
\end{itemize}

\begin{table}[htbp]
\centering
\small
\renewcommand{\arraystretch}{1.3}
\caption{Model configurations and hyperparameters.}
\label{tab:model_configs}
\begin{tabularx}{\columnwidth}{@{} p{0.28\columnwidth} >{\raggedright\arraybackslash}X @{}} 
\toprule
\textbf{Model} & \textbf{Configuration} \\
\midrule
\textbf{XGBoost} & \texttt{scale\_pos\_weight}=balanced, \texttt{eval\_metric}=logloss, \texttt{n\_jobs}=-1, \texttt{random\_state}=42 \\
\textbf{MLP} & \texttt{layers}=(100, 50), \texttt{activation}=relu, \texttt{solver}=adam, \texttt{max\_iter}=500 \\
\textbf{BERT} & Base: \texttt{bert-base-uncased}, opt: AdamW (lr=2e-5), batch: 32, epochs: 4 \\
\textbf{InceptionV3} & Pretrained: ImageNet, opt: Adam (lr=1e-4), loss: BCE, batch: 32 \\
\textbf{CLIP} & Backbone: ViT-B/32, LR: 1e-6 (encoder) / 1e-4 (head), batch: 64 \\
\bottomrule
\end{tabularx}
\end{table}

\subsection{Evaluation Metrics} \label{subsec:eval_metrics}
We report results from a \textbf{five-fold stratified cross-validation} on the training set to obtain a stable performance estimate. Stratification preserves the class distribution across folds, which is critical given the strong imbalance. Our primary metric is the \textbf{Area Under the Precision-Recall Curve (PR AUC)}, which is well suited to unbalanced classification tasks where the positive class is of main interest\cite{Davis2006}. We also report ROC AUC and F1 score to provide a fuller view of model sensitivity and specificity. For each time window, we then evaluate the best-performing model on the chronologically held-out test set to measure generalisation performance on future data.

\section{Experiments and Results} \label{sec:experiments_results}

This section reports the predictive performance of our framework against linear baselines and end-to-end deep learning architectures. All results use the \textbf{strict chronological train-test split}, ensuring methodological rigour and simulating real-world forecasting conditions. We first present the main results across increasing time windows and show that our feature-based approach scales well. We then compare our models with deep learning baselines (BERT, InceptionV3, CLIP) to test the ``Content Ceiling''. Finally, we analyse the ``evidentiary transition'' in feature importance over time.

\begin{table}[htbp]
    \small 
    \centering
    \caption{Test set performance of XGBoost and MLP neural network baseline across time windows.}
    \label{tab:main_results}
    \resizebox{\columnwidth}{!}{
        \begin{tabular}{@{}llcccc@{}}
            \toprule
            Time Window & Model Type             & PR AUC & ROC AUC & F1 Score & Duration (s) \\
            (min)       &                        &        &         &          &              \\
            \midrule
            30          & MLP Neural Network     & 0.38  & 0.88   & 0.24    & 169.62       \\
            30          & XGBoost                & \textbf{0.43} & \textbf{0.92} & \textbf{0.44} & \textbf{31.76} \\
            \midrule
            60          & MLP Neural Network     & 0.48  & 0.90   & 0.39    & 156.32       \\
            60          & XGBoost                & \textbf{0.54} & \textbf{0.94} & \textbf{0.51} & \textbf{24.91} \\
            \midrule
            120         & MLP Neural Network     & 0.59  & 0.92   & 0.52    & 154.67       \\
            120         & XGBoost                & \textbf{0.65} & \textbf{0.95} & \textbf{0.57} & \textbf{25.50} \\
            \midrule
            180         & MLP Neural Network     & 0.64  & 0.93   & 0.60    & 167.87       \\
            180         & XGBoost                & \textbf{0.68} & \textbf{0.96} & \textbf{0.62} & \textbf{26.14} \\
            \midrule
            240         & MLP Neural Network     & 0.69  & 0.94   & 0.62    & 155.64       \\
            240         & XGBoost                & \textbf{0.72} & \textbf{0.96} & \textbf{0.66} & \textbf{26.08} \\
            \midrule
            300         & MLP Neural Network     & 0.72  & 0.95   & 0.64    & 156.94       \\
            300         & XGBoost                & \textbf{0.75} & \textbf{0.97} & \textbf{0.66} & \textbf{26.21} \\
            \midrule
            360         & MLP Neural Network     & 0.75  & 0.96   & 0.67    & 154.41       \\
            360         & XGBoost                & \textbf{0.77} & \textbf{0.97} & \textbf{0.69} & \textbf{26.24} \\
            \midrule
            420         & MLP Neural Network     & 0.77  & 0.96   & 0.69    & 155.23       \\
            420         & XGBoost                & \textbf{0.80} & \textbf{0.98} & \textbf{0.73} & \textbf{25.81} \\
            \bottomrule
        \end{tabular}
    }
\end{table}

\subsection{Main Results: Performance Across Time Windows} \label{subsec:main_results}
Table~\ref{tab:main_results} summarises the performance of our models (XGBoost and the MLP neural network baseline) across increasing observation windows. The results show a clear pattern: as more engagement data becomes available over time, the predictive power of all models increases.
\textbf{XGBoost} is the strongest performer in every time window and on all metrics, which highlights the effectiveness of gradient boosting for this task. It reaches a PR AUC of \textbf{0.43} within the first \textbf{30 minutes} and rises to \textbf{0.80} after \textbf{420 minutes}. The \textbf{MLP neural network} provides a solid deep learning baseline, yet XGBoost consistently outperforms it, indicating that tree-based models handle this tabular feature set more effectively. The duration column also shows that XGBoost is computationally efficient, delivering the best performance in the shortest training time.

\subsection{Comparative Modality Analysis} \label{subsec:ablation_comparative}
To isolate the drivers of virality and benchmark our feature-based approach, we conduct a comparative analysis that contrasts deep learning baselines with our framework's individual and combined modalities (Table~\ref{tab:dl_vs_feature}). For clarity, all results in Table~\ref{tab:dl_vs_feature} are reported for the 120-minute observation window, which provides a representative early-range snapshot of model behaviour.

\begin{table}[htbp]
\centering
\small
\caption{Performance comparison: deep learning vs. feature engineering at the 120-minute observation window.}
\label{tab:dl_vs_feature}
\begin{tabularx}{\columnwidth}{@{} l l >{\raggedright\arraybackslash}X c c @{}}
\toprule
\textbf{Model} & \textbf{Approach} & \textbf{Modality} & \textbf{PR AUC} & \textbf{ROC AUC} \\
\midrule
InceptionV3 & Fine-tuned & Visual & 0.06 & 0.56 \\
BERT & Fine-tuned & Textual  & 0.13 & 0.62 \\
CLIP & Fine-tuned & Vis+Text & 0.10 & 0.61 \\
XGBoost & Feature eng.  & Visual & 0.06 & 0.57 \\
XGBoost & Feature eng.  & Textual & 0.11 & 0.63 \\
XGBoost & Feature eng.  & Contextual & 0.07 & 0.59 \\
XGBoost & Feature eng.  & Network & 0.35 & 0.90 \\
XGBoost & Feature eng. & Temporal  & 0.54 & 0.87 \\
XGBoost & Feature eng. & Vis+Text & 0.12 & 0.64 \\
XGBoost & Feature eng. & Vis+Text+Cont & 0.13 & 0.67 \\
XGBoost & Feature eng. & Vis+Text+Cont+Net & 0.43 & 0.92 \\
\textbf{XGBoost} & \textbf{Feature eng.} & \textbf{All (inc. Temp+Net)} & \textbf{0.65} & \textbf{0.95} \\
\bottomrule
\end{tabularx}
\end{table}

The results reveal a distinct ``Content Ceiling''. Whether we use end-to-end deep learning (BERT, CLIP, InceptionV3) or LLM-derived static features (visual, textual, contextual), models that rely only on intrinsic content consistently plateau at a low PR AUC of approximately \textbf{0.13}. This shows that content alone, no matter how sophisticated the extraction, is a weak predictor of virality without social context. The parity between LLM-derived content features and deep baselines (BERT, InceptionV3, CLIP) shows that the LLM \textbf{captures the same content signal ceiling} while providing interpretable tags that support feature-importance analysis in a way that black-box deep models do not.

The performance gains come from structural signals. The \textbf{Network} modality alone (PR AUC \textbf{0.35}) nearly triples the performance of the best content models by leveraging author and community context. The \textbf{Temporal} modality is even more critical, with early engagement dynamics alone reaching a PR AUC of \textbf{0.54}. Our full multimodal XGBoost model (PR AUC \textbf{0.65}) combines these structural signals with static content and \textbf{outperforms the deep learning baselines}, while remaining interpretable and computationally efficient.

\subsection{Feature Importance Analysis Over Time}
\label{subsec:feature_importance}

We use SHAP values \cite{lundberg2017unified} to examine how the relative importance of each modality evolves over time and which concrete features drive predictions in the full multimodal XGBoost model.

\begin{figure}[htbp]
    \centering
    \includegraphics[width=\columnwidth]{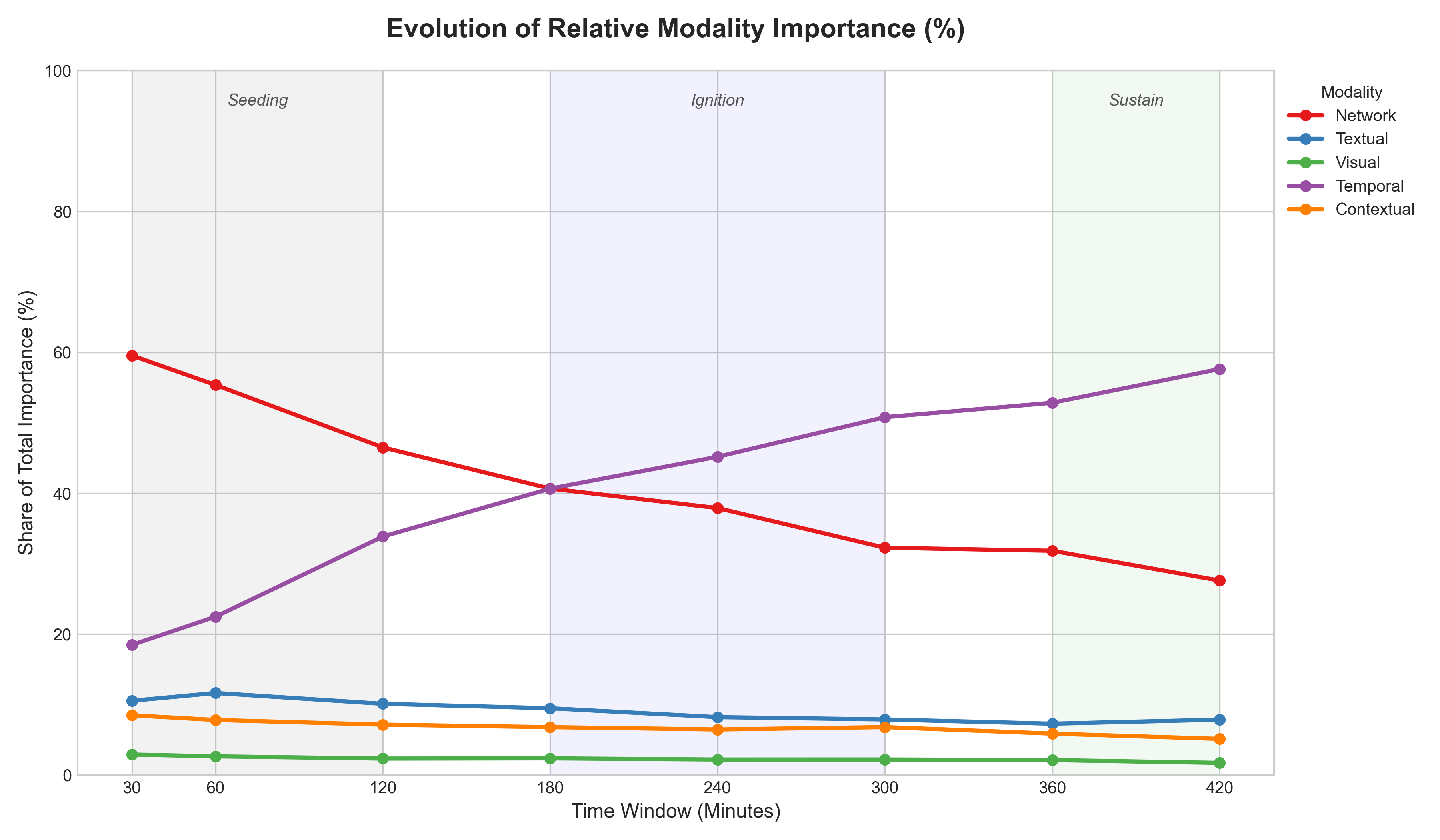}
    \caption{Evolution of feature modality importance based on SHAP values for XGBoost models trained at different time windows.}
    \label{fig:modality_importance}
\end{figure}

In the early seeding phase (\textbf{30} minutes), \textbf{Network} features dominate, contributing approximately \textbf{60\%} of the total predictive signal and highlighting the key role of community structure. As engagement data accumulates, a clear shift emerges. \textbf{Temporal} features rise from about \textbf{18\%} at \textbf{30} minutes and overtake Network features by the \textbf{180}-minute mark, where both are close to \textbf{40\%}. By the sustain phase (\textbf{420} minutes), Temporal dynamics become the primary driver (about \textbf{58\%}), while Network importance contracts to around \textbf{28\%}. Throughout this transition, intrinsic content features remain secondary: \textbf{Textual} and \textbf{Contextual} importance gradually decline (from roughly \textbf{11\%} and \textbf{8\%} to about \textbf{8\%} and \textbf{5\%}, respectively), and \textbf{Visual} features remain consistently marginal (below \textbf{3\%}).

\begin{table}[h!]
\centering
\caption{Spearman's rank correlation of proportional SHAP importance vs.\ time ($N = 8$).}
\label{tab:shap_spearman}
\begin{tabular}{@{}lcc@{}}
\toprule
\textbf{Modality} & \textbf{Spearman's $\rho$} & \textbf{$p$-value} \\
\midrule
Temporal   & \textbf{1.0000}   & $< 0.001$ \\
Network    & \textbf{-1.0000}  & $< 0.001$ \\
Visual     & \textbf{-1.0000}  & $< 0.001$ \\
Textual    & -0.9524           & $0.0003$ \\
Contextual & -0.9286           & $0.0009$ \\
\bottomrule
\end{tabular}
\end{table}

\begin{figure}[htbp]
    \centering
    \includegraphics[width=\columnwidth]{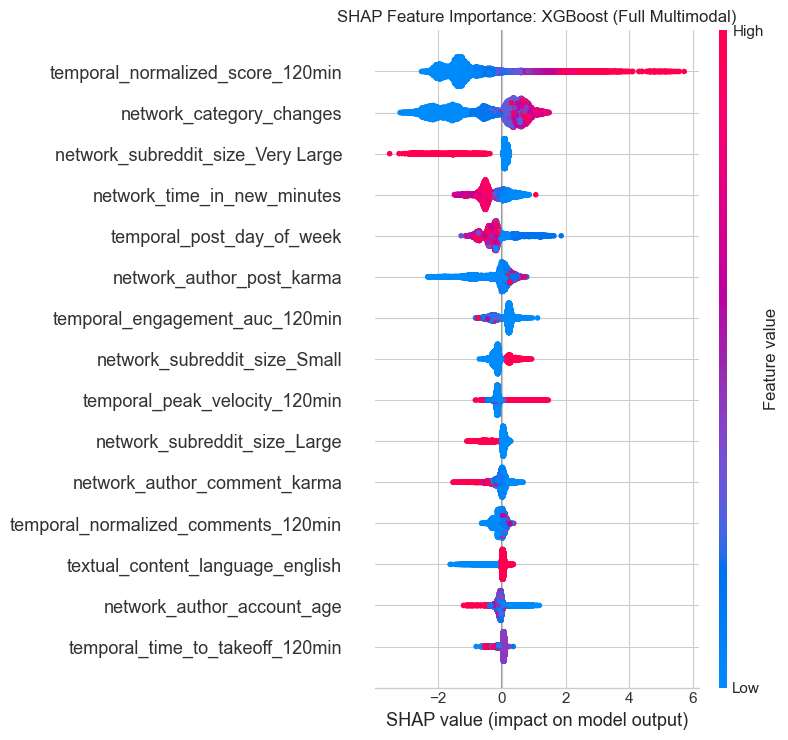}
    \caption{SHAP feature importance summary for the full multimodal XGBoost model at 120 minutes.}
    \label{fig:shap_full_multimodal}
\end{figure}

Spearman rank correlations on modality-level SHAP importance (Table~\ref{tab:shap_spearman}) confirm this pattern, with Temporal importance increasing monotonically over time while Network, Visual, Textual, and Contextual importance all decrease, which supports our \textbf{``evidentiary transition''} hypothesis. To understand which concrete signals drive this shift, we also inspect feature-level SHAP values for the full multimodal model at the representative \textbf{120}-minute window. Temporal engagement dominates. Normalised score at 120 minutes (\texttt{temporal\_normalized\_score\_120min}), engagement AUC, and peak velocity have the largest positive impact, with high early, community-normalised engagement pushing predictions towards the viral class. Network context provides strong structural priors through category transitions, time spent in \emph{New}, subreddit size bands, and author karma. Intrinsic content features such as language appear but remain secondary in this ranking, which reinforces the view that content sharpens, but does not replace, the structural and temporal evidence on which the model primarily relies  (Figure~\ref{fig:shap_full_multimodal}).

\section{Discussion, Limitations and Conclusion} 
\label{sec:discussion_conclusion}
\subsection{Discussion}

We asked whether we can predict the virality of cross-lingual Reddit memes early, which modalities matter most, and how their importance changes over time. The results in Section~\ref{sec:experiments_results} show that early prediction is feasible, that structural signals are more informative than intrinsic content, and that the evidential basis of prediction shifts over a meme's lifespan. A further contribution is our \emph{data-driven definition of virality}, which replaces simple volume counts and arbitrary thresholds with a principled target variable.

Using only \textbf{30 minutes} of engagement, the multimodal XGBoost model achieves a PR AUC of \textbf{0.43} and an ROC AUC of \textbf{0.92}, despite only \textbf{4.73\%} of posts being labelled viral (Table~\ref{tab:main_results}). Performance then increases to a PR AUC of \textbf{0.80} and an ROC AUC of \textbf{0.98} at \textbf{420 minutes}, which indicates that early trajectories carry substantial predictive signal and that the feature engineering pipeline is effective. Methodologically, our data-driven target variable addresses a key gap in prior work, where virality is usually defined with raw counts or arbitrary top-$x\%$ cut-offs that ignore community context and temporal dynamics (Section~\ref{subsec:target_definition}). The modality comparison shows a clear ``Content Ceiling'' (Table~\ref{tab:dl_vs_feature}). Content-only models, whether end-to-end deep learning (BERT, CLIP, InceptionV3) or feature-based (visual, textual, contextual), plateau at a PR AUC of around \textbf{0.13}. In contrast, Network features alone reach \textbf{0.35}, Temporal features alone reach \textbf{0.54}, and the full multimodal XGBoost model reaches \textbf{0.65}. Content patterns from the EDA (for example, video format, minimalist layouts, human faces, nostalgic or positive sentiment) are informative, but they are not sufficient to predict virality without structural and temporal context.

The SHAP analysis over time makes the ``evidentiary transition'' explicit (Figure~\ref{fig:modality_importance}). At \textbf{30 minutes}, Network features contribute about \textbf{60\%} of the top predictors and Temporal features about \textbf{18\%}. By \textbf{420 minutes}, Temporal features dominate at roughly \textbf{58\%}, while Network features fall to about \textbf{28\%}. Spearman correlations confirm that Temporal importance increases with time and that Network and Visual importance decrease (Table~\ref{tab:shap_spearman}). Early predictions are therefore driven by structural priors (author and community context), while accumulating engagement evidence (velocity, acceleration, take-off time) becomes decisively more informative later on. Virality thus appears as a dynamic, path-dependent process rather than a fixed property of content.

For platforms, these findings suggest that structural and temporal features can support early warning systems for recommendation, trend detection, and proactive moderation, especially where offensive or sensitive memes are more likely to become viral. The ``Content Ceiling'' implies that content-only models are unlikely to change virality dynamics without changes to exposure mechanisms. Time-aware pipelines that emphasise structural priors in very early windows and temporal dynamics later on, combined with our data-driven target definition, offer a principled way to operationalise virality without arbitrary thresholds.

\subsection{Limitations and Future Work}
This work has several limitations. Our dataset focuses on meme-centric subreddits over a relatively short period, which may limit generalisability to other content types, platforms, and time scales. Memes have specific formats and audiences. Future research should test whether the \textbf{Content Ceiling} and \textbf{evidentiary transition} effects replicate for news, long-form discussion, or other media domains. The Hybrid Score (through its fixed-percentile pretraining target) and the K-Means threshold, while data driven, still rest on specific modelling choices and implicit cut offs. Alternative virality definitions based on survival analysis, hazard rates, or multi-level engagement bands could be explored to test robustness and to address questions about time to peak or sustained attention. LLM-derived contextual labels may also introduce biases in how offensiveness, cultural references, or target audiences are perceived. Comparing these labels with human-coded annotations is an important next step.

Finally, the models we present are \textbf{predictive, not causal}. They quantify associations between features and virality but do not identify which interventions would change outcomes. Future work could investigate causal or intervention-based frameworks that explicitly model exposure and amplification, and study how algorithmic ranking and human behaviour interact to generate viral cascades.

\subsection{Conclusion}

This study introduced three main components: (i) a time-series, cross-lingual dataset of Reddit memes, (ii) a data-driven definition of virality based on a Hybrid Score, and (iii) a multimodal framework for early virality prediction. Together, these tackle a central gap in the literature, where virality is often treated as a simple function of final engagement volume with arbitrary thresholds that ignore community size, dynamics, and temporal ordering.

Empirically, we show that virality can be predicted with substantial accuracy within minutes of posting, even under strong class imbalance. The best-performing model, a multimodal XGBoost classifier, achieves a PR AUC of \textbf{0.43} at \textbf{30 minutes} and \textbf{0.80} at \textbf{420 minutes}, clearly surpassing content-only baselines. The results demonstrate a \textbf{Content Ceiling} for intrinsic content and highlight the need for Network and Temporal features to move beyond it. The temporal SHAP analysis further reveals a measurable \textbf{evidentiary transition} from structural priors to realised engagement dynamics over a meme's lifespan.

Overall, we reframe virality as an outcome shaped by where and when content appears and how quickly users respond, rather than as an intrinsic property of the meme alone. Future work can extend this foundation to other platforms and content types, incorporate richer temporal and causal models, and examine how human and algorithmic decisions jointly determine which memes become truly viral.

\section*{Data and Code Availability} 
\textbf{https://github.com/sdogan13/Meme-Analysis-and-Virality-Prediction}.

\bibliographystyle{ACM-Reference-Format}
\bibliography{references}


\begin{thebibliography}{32}


\ifx \showCODEN    \undefined \def \showCODEN     #1{\unskip}     \fi
\ifx \showISBNx    \undefined \def \showISBNx     #1{\unskip}     \fi
\ifx \showISBNxiii \undefined \def \showISBNxiii  #1{\unskip}     \fi
\ifx \showISSN     \undefined \def \showISSN      #1{\unskip}     \fi
\ifx \showLCCN     \undefined \def \showLCCN      #1{\unskip}     \fi
\ifx \shownote     \undefined \def \shownote      #1{#1}          \fi
\ifx \showarticletitle \undefined \def \showarticletitle #1{#1}   \fi
\ifx \showURL      \undefined \def \showURL       {\relax}        \fi
\providecommand\bibfield[2]{#2}
\providecommand\bibinfo[2]{#2}
\providecommand\natexlab[1]{#1}
\providecommand\showeprint[2][]{arXiv:#2}

\bibitem[Ageev et~al\mbox{.}(2024)]%
        {ageev_cognitive_2024}
\bibfield{author}{\bibinfo{person}{Sergei Ageev}, \bibinfo{person}{Evgeny Pushkarev}, {and} \bibinfo{person}{Natalia Antonenko}.} \bibinfo{year}{2024}\natexlab{}.
\newblock \showarticletitle{Cognitive underpinnings of misperceptions in morphed humor}.
\newblock \bibinfo{journal}{\emph{Russian Journal of Linguistics}} (\bibinfo{date}{June} \bibinfo{year}{2024}).
\newblock
\href{https://doi.org/10.22363/2687-0088-36029}{doi:\nolinkurl{10.22363/2687-0088-36029}}


\bibitem[Barnes et~al\mbox{.}(2024)]%
        {barnes_topicality_2024}
\bibfield{author}{\bibinfo{person}{Kate Barnes}, \bibinfo{person}{Péter Juhász}, \bibinfo{person}{Marcell Nagy}, {and} \bibinfo{person}{Roland Molontay}.} \bibinfo{year}{2024}\natexlab{}.
\newblock \showarticletitle{Topicality boosts popularity: a comparative analysis of NYT articles and Reddit memes}.
\newblock \bibinfo{journal}{\emph{Social Network Analysis and Mining}}  \bibinfo{volume}{14} (\bibinfo{year}{2024}), \bibinfo{pages}{1--18}.
\newblock
\href{https://doi.org/10.1007/s13278-024-01272-3}{doi:\nolinkurl{10.1007/s13278-024-01272-3}}


\bibitem[Barnes et~al\mbox{.}(2020)]%
        {barnes_dank_2020}
\bibfield{author}{\bibinfo{person}{Kate Barnes}, \bibinfo{person}{Tiernon Riesenmy}, \bibinfo{person}{Minh~Duc Trinh}, \bibinfo{person}{Eli Lleshi}, \bibinfo{person}{Nóra Balogh}, {and} \bibinfo{person}{Roland Molontay}.} \bibinfo{year}{2020}\natexlab{}.
\newblock \showarticletitle{Dank or not? {Analyzing} and predicting the popularity of memes on {Reddit}}.
\newblock \bibinfo{journal}{\emph{Applied Network Science}}  \bibinfo{volume}{6} (\bibinfo{date}{Nov.} \bibinfo{year}{2020}).
\newblock
\href{https://doi.org/10.1007/s41109-021-00358-7}{doi:\nolinkurl{10.1007/s41109-021-00358-7}}


\bibitem[Chen et~al\mbox{.}(2019)]%
        {chen_npp_2019}
\bibfield{author}{\bibinfo{person}{Guandan Chen}, \bibinfo{person}{Qingchao Kong}, \bibinfo{person}{Nan Xu}, {and} \bibinfo{person}{W. Mao}.} \bibinfo{year}{2019}\natexlab{}.
\newblock \showarticletitle{{NPP}: {A} neural popularity prediction model for social media content}.
\newblock \bibinfo{journal}{\emph{Neurocomputing}}  \bibinfo{volume}{333} (\bibinfo{date}{March} \bibinfo{year}{2019}), \bibinfo{pages}{221--230}.
\newblock
\href{https://doi.org/10.1016/j.neucom.2018.12.039}{doi:\nolinkurl{10.1016/j.neucom.2018.12.039}}


\bibitem[Chen and Guestrin(2016)]%
        {chen2016xgboost}
\bibfield{author}{\bibinfo{person}{Tianqi Chen} {and} \bibinfo{person}{Carlos Guestrin}.} \bibinfo{year}{2016}\natexlab{}.
\newblock \showarticletitle{XGBoost: A Scalable Tree Boosting System}. In \bibinfo{booktitle}{\emph{Proceedings of the 22nd ACM SIGKDD International Conference on Knowledge Discovery and Data Mining (KDD)}}. \bibinfo{pages}{785--794}.
\newblock
\href{https://doi.org/10.1145/2939672.2939785}{doi:\nolinkurl{10.1145/2939672.2939785}}


\bibitem[Cheng et~al\mbox{.}(2024)]%
        {cheng_information_2024}
\bibfield{author}{\bibinfo{person}{Zhangtao Cheng}, \bibinfo{person}{Fan Zhou}, \bibinfo{person}{Xovee Xu}, \bibinfo{person}{Kunpeng Zhang}, \bibinfo{person}{Goce Trajcevski}, \bibinfo{person}{Ting Zhong}, {and} \bibinfo{person}{Phillp Yu}.} \bibinfo{year}{2024}\natexlab{}.
\newblock \showarticletitle{Information {Cascade} {Popularity} {Prediction} via {Probabilistic} {Diffusion}}.
\newblock \bibinfo{journal}{\emph{IEEE Transactions on Knowledge and Data Engineering}}  \bibinfo{volume}{36} (\bibinfo{date}{Dec.} \bibinfo{year}{2024}), \bibinfo{pages}{8541--8555}.
\newblock
\href{https://doi.org/10.1109/TKDE.2024.3465241}{doi:\nolinkurl{10.1109/TKDE.2024.3465241}}


\bibitem[Cieslewski et~al\mbox{.}(2022)]%
        {cieslewski_continuous-time_2022}
\bibfield{author}{\bibinfo{person}{Titus Cieslewski}, \bibinfo{person}{D. Scaramuzza}, {and} \bibinfo{person}{Giovanni Cioffi}.} \bibinfo{year}{2022}\natexlab{}.
\newblock \showarticletitle{Continuous-{Time} vs. {Discrete}-{Time} {Vision}-based {SLAM}: {A} {Comparative} {Study}}.
\newblock \bibinfo{journal}{\emph{IEEE Robotics and Automation Letters}}  \bibinfo{volume}{PP} (\bibinfo{date}{Feb.} \bibinfo{year}{2022}), \bibinfo{pages}{1--1}.
\newblock
\href{https://doi.org/10.1109/lra.2022.3143303}{doi:\nolinkurl{10.1109/lra.2022.3143303}}


\bibitem[Crispino et~al\mbox{.}(2023)]%
        {wang_agent_2023}
\bibfield{author}{\bibinfo{person}{Nicholas Crispino}, \bibinfo{person}{Kyle Montgomery}, \bibinfo{person}{Fankun Zeng}, \bibinfo{person}{Dawn Song}, {and} \bibinfo{person}{Chenguang Wang}.} \bibinfo{year}{2023}\natexlab{}.
\newblock \showarticletitle{Agent {Instructs} {Large} {Language} {Models} to be {General} {Zero}-{Shot} {Reasoners}}.
\newblock \bibinfo{journal}{\emph{ArXiv}}  \bibinfo{volume}{abs/2310.03710} (\bibinfo{date}{Oct.} \bibinfo{year}{2023}).
\newblock
\href{https://doi.org/10.48550/arXiv.2310.03710}{doi:\nolinkurl{10.48550/arXiv.2310.03710}}


\bibitem[Davis and Goadrich(2006)]%
        {Davis2006}
\bibfield{author}{\bibinfo{person}{Jesse Davis} {and} \bibinfo{person}{Mark Goadrich}.} \bibinfo{year}{2006}\natexlab{}.
\newblock \showarticletitle{The Relationship Between Precision-Recall and {ROC} Curves}. In \bibinfo{booktitle}{\emph{Proceedings of the 23rd International Conference on Machine Learning (ICML)}}. \bibinfo{publisher}{ACM}, \bibinfo{address}{Pittsburgh, PA}, \bibinfo{pages}{233--240}.
\newblock
\href{https://doi.org/10.1145/1143844.1143874}{doi:\nolinkurl{10.1145/1143844.1143874}}


\bibitem[De~Haan-Rietdijk et~al\mbox{.}(2017)]%
        {de_haan-rietdijk_discrete-_2017}
\bibfield{author}{\bibinfo{person}{S. De~Haan-Rietdijk}, \bibinfo{person}{M. Voelkle}, \bibinfo{person}{L. Keijsers}, {and} \bibinfo{person}{E. Hamaker}.} \bibinfo{year}{2017}\natexlab{}.
\newblock \showarticletitle{Discrete- vs. {Continuous}-{Time} {Modeling} of {Unequally} {Spaced} {Experience} {Sampling} {Method} {Data}}.
\newblock \bibinfo{journal}{\emph{Frontiers in Psychology}}  \bibinfo{volume}{8} (\bibinfo{date}{Oct.} \bibinfo{year}{2017}).
\newblock
\href{https://doi.org/10.3389/fpsyg.2017.01849}{doi:\nolinkurl{10.3389/fpsyg.2017.01849}}


\bibitem[Denisova(2020)]%
        {denisova_how_2020}
\bibfield{author}{\bibinfo{person}{Anastasia Denisova}.} \bibinfo{year}{2020}\natexlab{}.
\newblock \showarticletitle{How to {Define} ‘{Viral}’ for {Media} {Studies}?}
\newblock \bibinfo{journal}{\emph{Westminster Papers in Communication and Culture}} (\bibinfo{date}{March} \bibinfo{year}{2020}).
\newblock
\href{https://doi.org/10.16997/wpcc.375}{doi:\nolinkurl{10.16997/wpcc.375}}


\bibitem[Devlin et~al\mbox{.}(2019)]%
        {devlin2019bert}
\bibfield{author}{\bibinfo{person}{Jacob Devlin}, \bibinfo{person}{Ming{-}Wei Chang}, \bibinfo{person}{Kenton Lee}, {and} \bibinfo{person}{Kristina Toutanova}.} \bibinfo{year}{2019}\natexlab{}.
\newblock \showarticletitle{BERT: Pre-training of Deep Bidirectional Transformers for Language Understanding}. In \bibinfo{booktitle}{\emph{Proceedings of NAACL-HLT 2019}}.
\newblock
\href{https://doi.org/10.18653/v1/N19-1423}{doi:\nolinkurl{10.18653/v1/N19-1423}}


\bibitem[El-amrany et~al\mbox{.}(2025)]%
        {elamrany2025redditv}
\bibfield{author}{\bibinfo{person}{Samir El-amrany}, \bibinfo{person}{Matthias~R. Brust}, \bibinfo{person}{Salima Lamsiyah}, {and} \bibinfo{person}{Pascal Bouvry}.} \bibinfo{year}{2025}\natexlab{}.
\newblock \showarticletitle{Reddit-V: A Virality Prediction Dataset and Zero-Shot Evaluation with Large Language Models}. In \bibinfo{booktitle}{\emph{Proceedings of the 15th International Conference on Recent Advances in Natural Language Processing -- Natural Language Processing in the Generative AI era}}. \bibinfo{publisher}{INCOMA Ltd.}, \bibinfo{address}{Varna, Bulgaria}, \bibinfo{pages}{334--341}.
\newblock
\href{https://doi.org/2025.ranlp-1.41}{doi:\nolinkurl{2025.ranlp-1.41}}


\bibitem[{Gemini Team} and {Google}(2023)]%
        {GoogleGemini}
\bibfield{author}{\bibinfo{person}{{Gemini Team}} {and} \bibinfo{person}{{Google}}.} \bibinfo{year}{2023}\natexlab{}.
\newblock \bibinfo{title}{Gemini: A Family of Highly Capable Multimodal Models}.
\newblock \bibinfo{howpublished}{arXiv preprint arXiv:2312.11805}.
\newblock
\urldef\tempurl%
\url{https://arxiv.org/abs/2312.11805}
\showURL{%
\tempurl}
\newblock
\shownote{Accessed: May 1, 2025}.


\bibitem[Gleeson et~al\mbox{.}(2016)]%
        {gleeson_limitations_2016}
\bibfield{author}{\bibinfo{person}{J. Gleeson}, \bibinfo{person}{S. Melnik}, {and} \bibinfo{person}{Peter Fennell}.} \bibinfo{year}{2016}\natexlab{}.
\newblock \showarticletitle{Limitations of discrete-time approaches to continuous-time contagion dynamics}.
\newblock \bibinfo{journal}{\emph{Physical Review. E}}  \bibinfo{volume}{94} (\bibinfo{date}{Feb.} \bibinfo{year}{2016}).
\newblock
\href{https://doi.org/10.1103/PhysRevE.94.052125}{doi:\nolinkurl{10.1103/PhysRevE.94.052125}}


\bibitem[Goel et~al\mbox{.}(2015)]%
        {goel_structural_2015}
\bibfield{author}{\bibinfo{person}{Sharad Goel}, \bibinfo{person}{Ashton Anderson}, \bibinfo{person}{Jake Hofman}, {and} \bibinfo{person}{D. Watts}.} \bibinfo{year}{2015}\natexlab{}.
\newblock \showarticletitle{The {Structural} {Virality} of {Online} {Diffusion}}.
\newblock \bibinfo{journal}{\emph{Manag. Sci.}}  \bibinfo{volume}{62} (\bibinfo{date}{July} \bibinfo{year}{2015}), \bibinfo{pages}{180--196}.
\newblock
\href{https://doi.org/10.1287/mnsc.2015.2158}{doi:\nolinkurl{10.1287/mnsc.2015.2158}}


\bibitem[Guseynova et~al\mbox{.}(2022)]%
        {guseynova_reflection_2022}
\bibfield{author}{\bibinfo{person}{L. Guseynova}, \bibinfo{person}{Natalia Dugalich}, \bibinfo{person}{O. Lomakina}, \bibinfo{person}{Natalia~Yu. Neliubova}, {and} \bibinfo{person}{Y. Ebzeeva}.} \bibinfo{year}{2022}\natexlab{}.
\newblock \showarticletitle{The {Reflection} of the {Socio}-{Cultural} {Context} in {Russian}, {French} and {Azerbaijani} {Internet} {Memes}}.
\newblock \bibinfo{journal}{\emph{RUDN Journal of Language Studies, Semiotics and Semantics}} (\bibinfo{date}{Dec.} \bibinfo{year}{2022}).
\newblock
\href{https://doi.org/10.22363/2313-2299-2022-13-4-1020-1043}{doi:\nolinkurl{10.22363/2313-2299-2022-13-4-1020-1043}}


\bibitem[Ling et~al\mbox{.}(2021)]%
        {ling_dissecting_2021}
\bibfield{author}{\bibinfo{person}{Chen Ling}, \bibinfo{person}{Ihab AbuHilal}, \bibinfo{person}{Jeremy Blackburn}, \bibinfo{person}{Emiliano De~Cristofaro}, \bibinfo{person}{Savvas Zannettou}, {and} \bibinfo{person}{G. Stringhini}.} \bibinfo{year}{2021}\natexlab{}.
\newblock \showarticletitle{Dissecting the {Meme} {Magic}: {Understanding} {Indicators} of {Virality} in {Image} {Memes}}.
\newblock \bibinfo{journal}{\emph{Proceedings of the ACM on Human-Computer Interaction}}  \bibinfo{volume}{5} (\bibinfo{date}{Jan.} \bibinfo{year}{2021}), \bibinfo{pages}{1--24}.
\newblock
\href{https://doi.org/10.1145/3449155}{doi:\nolinkurl{10.1145/3449155}}


\bibitem[Liu(2024)]%
        {liu_case-based_2024}
\bibfield{author}{\bibinfo{person}{Bingqing Liu}.} \bibinfo{year}{2024}\natexlab{}.
\newblock \showarticletitle{A {Case}-{Based} {Reasoning} and {Explaining} {Model} for {Temporal} {Point} {Process}}.
\newblock  (\bibinfo{year}{2024}), \bibinfo{pages}{127--142}.
\newblock
\href{https://doi.org/10.1007/978-3-031-63646-2_9}{doi:\nolinkurl{10.1007/978-3-031-63646-2_9}}


\bibitem[Lundberg and Lee(2017)]%
        {lundberg2017unified}
\bibfield{author}{\bibinfo{person}{Scott~M. Lundberg} {and} \bibinfo{person}{Su-In Lee}.} \bibinfo{year}{2017}\natexlab{}.
\newblock \showarticletitle{A Unified Approach to Interpreting Model Predictions}. In \bibinfo{booktitle}{\emph{Advances in Neural Information Processing Systems (NeurIPS 2017)}}, Vol.~\bibinfo{volume}{30}. \bibinfo{pages}{4765--4774}.
\newblock
\href{https://doi.org/10.48550/arXiv.1705.07874}{doi:\nolinkurl{10.48550/arXiv.1705.07874}}


\bibitem[MacQueen(1967)]%
        {MacQueen1967}
\bibfield{author}{\bibinfo{person}{J. MacQueen}.} \bibinfo{year}{1967}\natexlab{}.
\newblock \showarticletitle{Some methods for classification and analysis of multivariate observations}. In \bibinfo{booktitle}{\emph{Proceedings of the Fifth Berkeley Symposium on Mathematical Statistics and Probability, Volume 1: Statistics}}. \bibinfo{publisher}{University of California Press}, \bibinfo{pages}{281--297}.
\newblock


\bibitem[Mills(2012)]%
        {mills_virality_2012}
\bibfield{author}{\bibinfo{person}{Adam Mills}.} \bibinfo{year}{2012}\natexlab{}.
\newblock \showarticletitle{Virality in social media: the {SPIN} {Framework}}.
\newblock \bibinfo{journal}{\emph{Journal of Public Affairs}}  \bibinfo{volume}{12} (\bibinfo{date}{May} \bibinfo{year}{2012}), \bibinfo{pages}{162--169}.
\newblock
\href{https://doi.org/10.1002/PA.1418}{doi:\nolinkurl{10.1002/PA.1418}}


\bibitem[Pedregosa et~al\mbox{.}(2011)]%
        {pedregosa2011sklearn}
\bibfield{author}{\bibinfo{person}{Fabian Pedregosa}, \bibinfo{person}{Ga{\"e}l Varoquaux}, \bibinfo{person}{Alexandre Gramfort}, \bibinfo{person}{Vincent Michel}, \bibinfo{person}{Bertrand Thirion}, \bibinfo{person}{Olivier Grisel}, \bibinfo{person}{Mathieu Blondel}, \bibinfo{person}{Peter Prettenhofer}, \bibinfo{person}{Ron Weiss}, \bibinfo{person}{Vincent Dubourg}, \bibinfo{person}{Jake Vanderplas}, \bibinfo{person}{Alexandre Passos}, \bibinfo{person}{David Cournapeau}, \bibinfo{person}{Matthieu Brucher}, \bibinfo{person}{Matthieu Perrot}, {and} \bibinfo{person}{{\'E}douard Duchesnay}.} \bibinfo{year}{2011}\natexlab{}.
\newblock \showarticletitle{Scikit-learn: Machine Learning in Python}.
\newblock \bibinfo{journal}{\emph{Journal of Machine Learning Research}}  \bibinfo{volume}{12} (\bibinfo{year}{2011}), \bibinfo{pages}{2825--2830}.
\newblock
\href{https://doi.org/10.48550/arXiv.1201.0490}{doi:\nolinkurl{10.48550/arXiv.1201.0490}}


\bibitem[Radford et~al\mbox{.}(2021)]%
        {Radford2021CLIP}
\bibfield{author}{\bibinfo{person}{Alec Radford}, \bibinfo{person}{Jong~Wook Kim}, \bibinfo{person}{Chris Hallacy}, \bibinfo{person}{Aditya Ramesh}, \bibinfo{person}{Gabriel Goh}, \bibinfo{person}{Sandhini Agarwal}, \bibinfo{person}{Girish Sastry}, \bibinfo{person}{Amanda Askell}, \bibinfo{person}{Pamela Mishkin}, \bibinfo{person}{Jack Clark}, \bibinfo{person}{Gretchen Krueger}, {and} \bibinfo{person}{Ilya Sutskever}.} \bibinfo{year}{2021}\natexlab{}.
\newblock \showarticletitle{Learning Transferable Visual Models From Natural Language Supervision}. In \bibinfo{booktitle}{\emph{Proceedings of the 38th International Conference on Machine Learning (ICML 2021)}}, Vol.~\bibinfo{volume}{139}. \bibinfo{publisher}{PMLR}, \bibinfo{pages}{8748--8763}.
\newblock
\href{https://doi.org/10.48550/arXiv.2103.00020}{doi:\nolinkurl{10.48550/arXiv.2103.00020}}


\bibitem[Rumelhart et~al\mbox{.}(1986)]%
        {Rumelhart1986}
\bibfield{author}{\bibinfo{person}{David~E. Rumelhart}, \bibinfo{person}{Geoffrey~E. Hinton}, {and} \bibinfo{person}{Ronald~J. Williams}.} \bibinfo{year}{1986}\natexlab{}.
\newblock \showarticletitle{Learning representations by back-propagating errors}.
\newblock \bibinfo{journal}{\emph{Nature}} \bibinfo{volume}{323}, \bibinfo{number}{6088} (\bibinfo{year}{1986}), \bibinfo{pages}{533--536}.
\newblock
\href{https://doi.org/10.1038/323533a0}{doi:\nolinkurl{10.1038/323533a0}}


\bibitem[Sah and Jordan(2025)]%
        {sah2025decoding}
\bibfield{author}{\bibinfo{person}{Tanmay Sah} {and} \bibinfo{person}{Kayden Jordan}.} \bibinfo{year}{2025}\natexlab{}.
\newblock \showarticletitle{Decoding reddit memes virality}.
\newblock \bibinfo{journal}{\emph{International Journal of Data Science and Analytics}} \bibinfo{volume}{20}, \bibinfo{number}{6} (\bibinfo{date}{Apr} \bibinfo{year}{2025}), \bibinfo{pages}{5321--5336}.
\newblock
\href{https://doi.org/10.1007/s41060-025-00772-5}{doi:\nolinkurl{10.1007/s41060-025-00772-5}}


\bibitem[Shifman(2014)]%
        {shifman_cultural_2014}
\bibfield{author}{\bibinfo{person}{L. Shifman}.} \bibinfo{year}{2014}\natexlab{}.
\newblock \showarticletitle{The {Cultural} {Logic} of {Photo}-{Based} {Meme} {Genres}}.
\newblock \bibinfo{journal}{\emph{Journal of Visual Culture}}  \bibinfo{volume}{13} (\bibinfo{date}{Dec.} \bibinfo{year}{2014}), \bibinfo{pages}{340--358}.
\newblock
\href{https://doi.org/10.1177/1470412914546577}{doi:\nolinkurl{10.1177/1470412914546577}}


\bibitem[Szegedy et~al\mbox{.}(2016)]%
        {szegedy2016rethinking}
\bibfield{author}{\bibinfo{person}{Christian Szegedy}, \bibinfo{person}{Wei Liu}, \bibinfo{person}{Yangqing Jia}, \bibinfo{person}{Pierre Sermanet}, \bibinfo{person}{Scott Reed}, \bibinfo{person}{Dongdong Zhu}, \bibinfo{person}{Hongwei Zhang}, {and} \bibinfo{person}{Zhengyou Zhang}.} \bibinfo{year}{2016}\natexlab{}.
\newblock \showarticletitle{Rethinking the Inception Architecture for Computer Vision}. In \bibinfo{booktitle}{\emph{Proceedings of the IEEE Conference on Computer Vision and Pattern Recognition (CVPR)}}.
\newblock
\href{https://doi.org/10.1109/CVPR.2016.308}{doi:\nolinkurl{10.1109/CVPR.2016.308}}


\bibitem[Weng et~al\mbox{.}(2013)]%
        {weng_virality_2013}
\bibfield{author}{\bibinfo{person}{L. Weng}, \bibinfo{person}{F. Menczer}, {and} \bibinfo{person}{Yong-Yeol Ahn}.} \bibinfo{year}{2013}\natexlab{}.
\newblock \showarticletitle{Virality {Prediction} and {Community} {Structure} in {Social} {Networks}}.
\newblock \bibinfo{journal}{\emph{Scientific Reports}}  \bibinfo{volume}{3} (\bibinfo{date}{June} \bibinfo{year}{2013}).
\newblock
\href{https://doi.org/10.1038/srep02522}{doi:\nolinkurl{10.1038/srep02522}}


\bibitem[Xu et~al\mbox{.}(2023)]%
        {xu_casflow_2023}
\bibfield{author}{\bibinfo{person}{Xovee Xu}, \bibinfo{person}{Fan Zhou}, \bibinfo{person}{Kunpeng Zhang}, \bibinfo{person}{Siyuan Liu}, {and} \bibinfo{person}{Goce Trajcevski}.} \bibinfo{year}{2023}\natexlab{}.
\newblock \showarticletitle{{CasFlow}: {Exploring} {Hierarchical} {Structures} and {Propagation} {Uncertainty} for {Cascade} {Prediction}}.
\newblock \bibinfo{journal}{\emph{IEEE Transactions on Knowledge and Data Engineering}}  \bibinfo{volume}{35} (\bibinfo{date}{April} \bibinfo{year}{2023}), \bibinfo{pages}{3484--3499}.
\newblock
\href{https://doi.org/10.1109/TKDE.2021.3126475}{doi:\nolinkurl{10.1109/TKDE.2021.3126475}}


\bibitem[Zhang et~al\mbox{.}(2025)]%
        {zhang_casformer_2025}
\bibfield{author}{\bibinfo{person}{Ji Zhang}, \bibinfo{person}{Zhao Li}, \bibinfo{person}{Biao Wang}, {and} \bibinfo{person}{Zenghui Xu}.} \bibinfo{year}{2025}\natexlab{}.
\newblock \showarticletitle{Casformer: {Information} {Popularity} {Prediction} {With} {Adaptive} {Cascade} {Sampling} and {Graph} {Transformer} in {Social} {Networks}}.
\newblock \bibinfo{journal}{\emph{IEEE Transactions on Big Data}} (\bibinfo{date}{Jan.} \bibinfo{year}{2025}).
\newblock
\href{https://doi.org/10.1109/tbdata.2024.3524839}{doi:\nolinkurl{10.1109/tbdata.2024.3524839}}


\bibitem[Zhang et~al\mbox{.}(2021)]%
        {zhang_survey_2021}
\bibfield{author}{\bibinfo{person}{Kunpeng Zhang}, \bibinfo{person}{Fan Zhou}, \bibinfo{person}{Goce Trajcevski}, {and} \bibinfo{person}{Xovee Xu}.} \bibinfo{year}{2021}\natexlab{}.
\newblock \showarticletitle{A {Survey} of {Information} {Cascade} {Analysis}: {Models}, {Predictions}, and {Recent} {Advances}}.
\newblock \bibinfo{journal}{\emph{ACM Computing Surveys (CSUR)}}  \bibinfo{volume}{54(2)} (\bibinfo{date}{March} \bibinfo{year}{2021}), \bibinfo{pages}{1--36}.
\newblock
\href{https://doi.org/10.1145/3433000}{doi:\nolinkurl{10.1145/3433000}}


\end{thebibliography}

\clearpage 
\appendix
\section{Appendix}
\label{sec:appendix}
\subsection{Exploratory Data Analysis (EDA)} \label{subsec:appendix_virality_plots}

\begin{figure}[h!]
  \centering
  \includegraphics[width=\columnwidth]{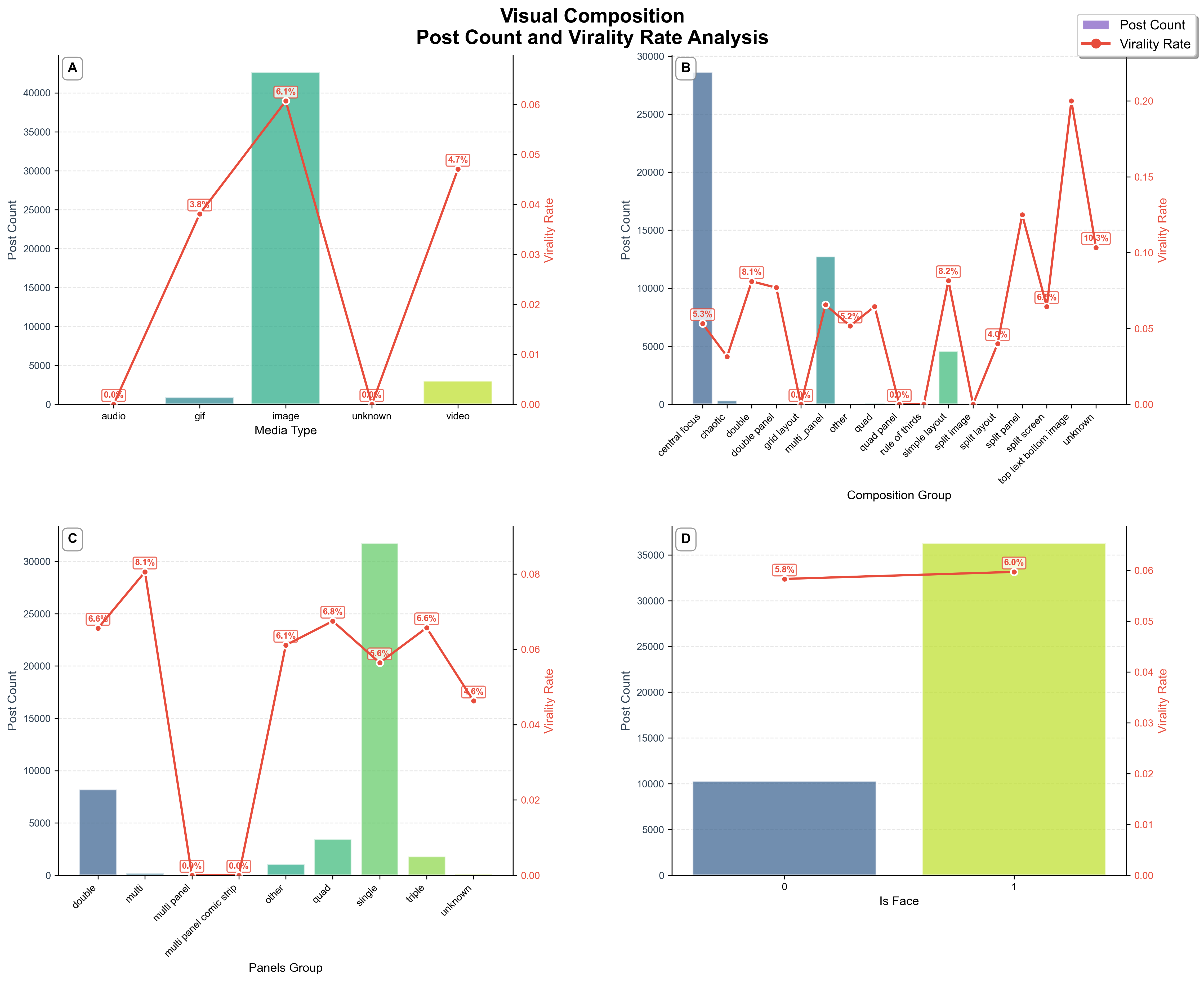}
  \caption{Virality rates for core visual elements. The subplots analyse virality based on (A) media type, (B) visual composition, (C) panel structure, and (D) the presence of a face.}
  \label{fig:visual_comps}
\end{figure}

\begin{figure}[htbp]
  \centering
  \includegraphics[width=\columnwidth]{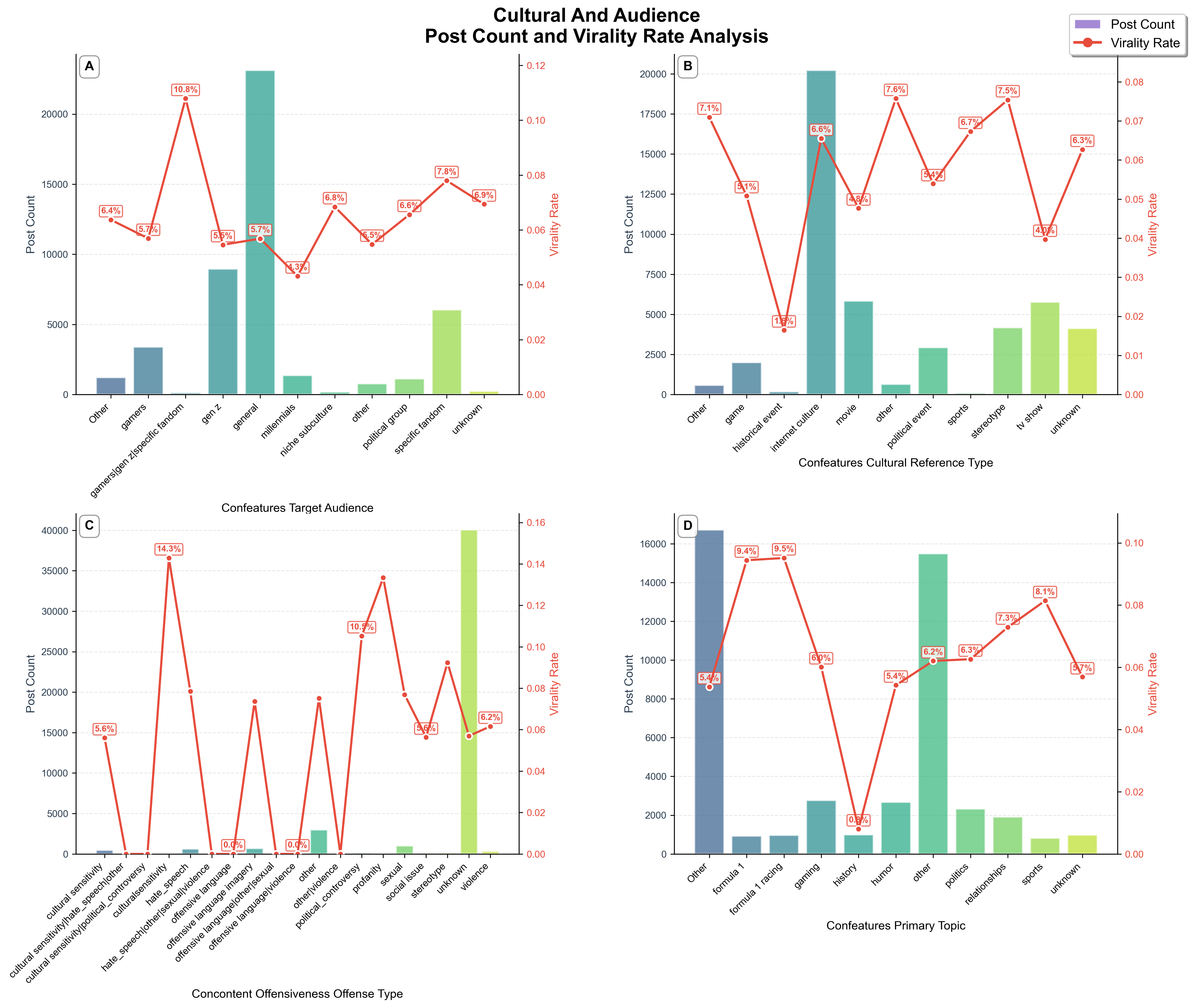}
  \caption{Virality rate distribution for cultural and audience-specific features. Subplots detail virality based on (A) content offensiveness, (B) audience target, (C) offence type, and (D) primary topic.}
  \label{fig:cultural_audience}
\end{figure}

\begin{figure}[htbp]
  \centering
  \includegraphics[width=\columnwidth]{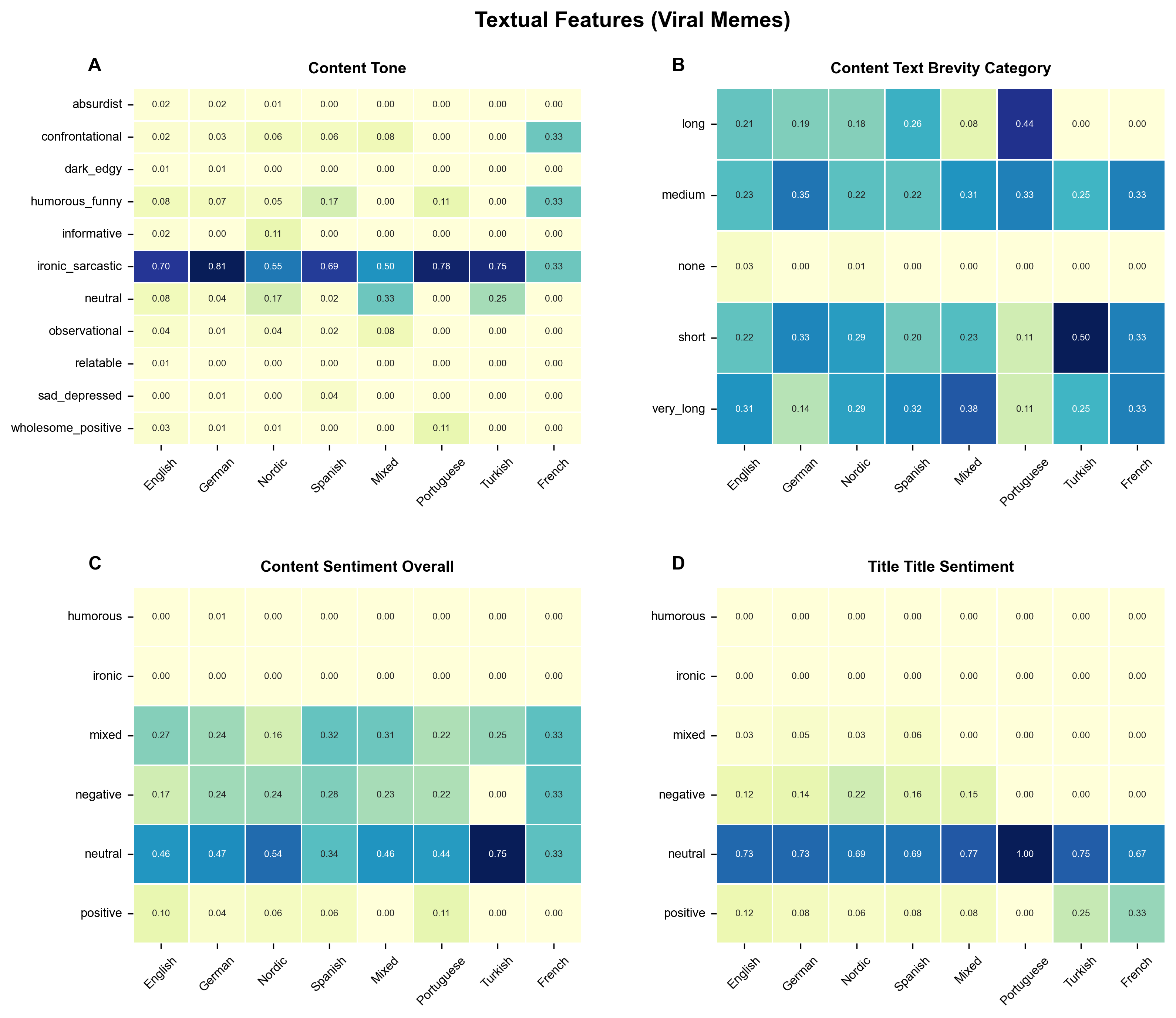}
  \caption{Cross-lingual distribution of textual features by language group (sentiment, tone and brevity.)}
  \label{fig:textual_crosslingual}
\end{figure}

\begin{figure}[htbp]
  \centering
  \includegraphics[width=\columnwidth]{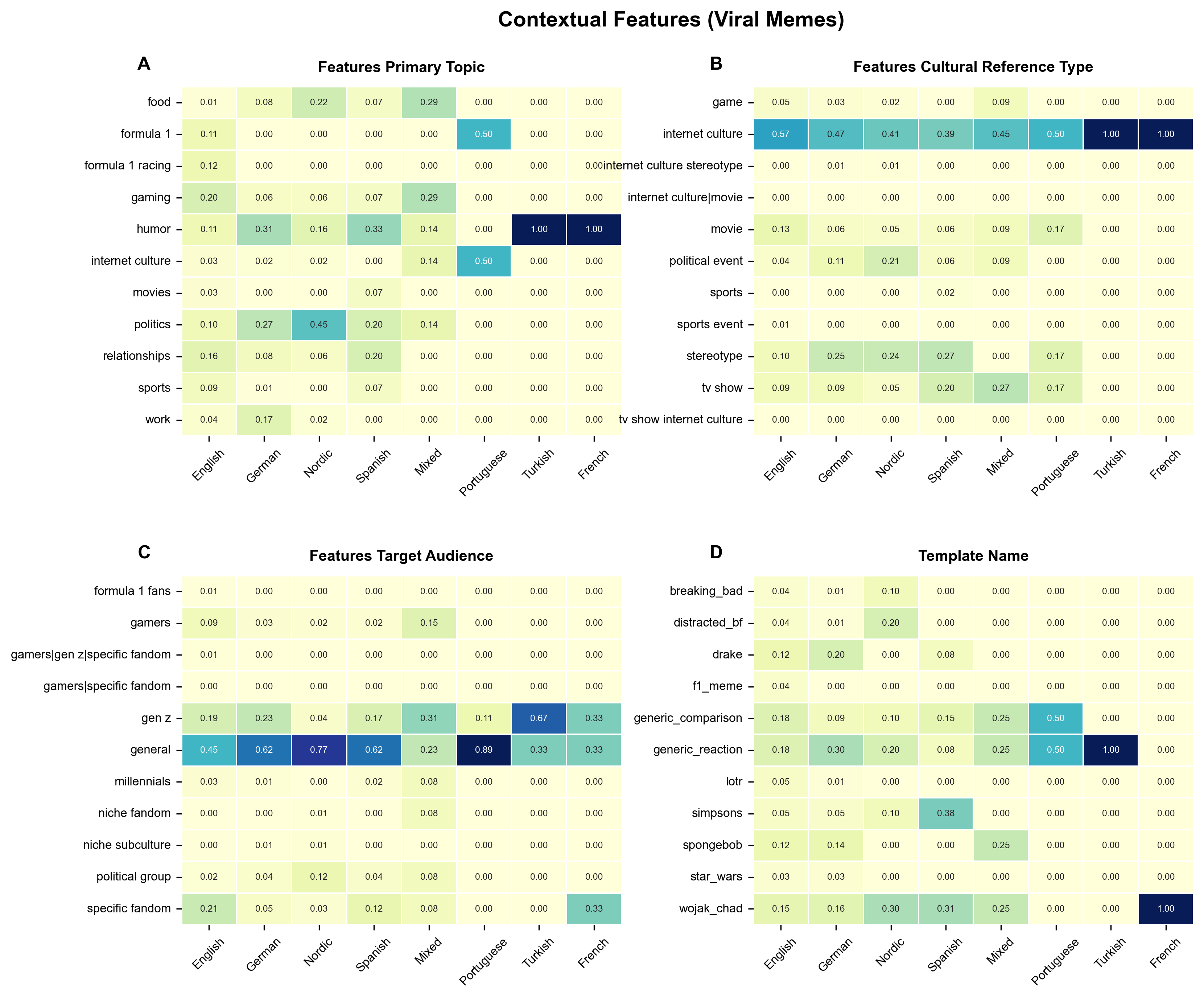}
  \caption{Cross-lingual distribution of contextual features by language group (template name, cultural reference type, target audience, and topic).}
  \label{fig:contextual_crosslingual}
\end{figure}

\end{document}